\documentclass[11pt,a4paper]{article}
\usepackage[hyperref]{acl2020}
\usepackage{times}
\usepackage{latexsym}

\usepackage{graphicx}
\graphicspath{ {./figures/} }
\usepackage{lipsum}
\usepackage{amsmath}
\usepackage{amsfonts}
\usepackage{subcaption}
\usepackage{multirow}

\usepackage{microtype}

\aclfinalcopy 
\def\aclpaperid{1096} 

\title{Neural Generation of Dialogue Response Timings}

\author{Matthew Roddy and Naomi Harte \\
  ADAPT Centre, School of Engineering \\
  Trinity College Dublin, Ireland \\
  \texttt{\{roddym,nharte\}@tcd.ie}}

\date{}

\begin{document}
\maketitle
\begin{abstract}
The timings of spoken response offsets in human dialogue have been shown to vary based on contextual elements of the dialogue. We propose neural models that simulate the distributions of these response offsets, taking into account the response turn as well as the preceding turn. The models are designed to be integrated into the pipeline of an incremental spoken dialogue system (SDS). We evaluate our models using offline experiments as well as human listening tests. We show that human listeners consider certain response timings to be more natural based on the dialogue context. The introduction of these models into SDS pipelines could increase the perceived naturalness of interactions.\footnote{\ Our code is available at \url{https://github.com/mattroddy/RTNets}.}
\end{abstract}

\section{Introduction \label{sec:Overview}}

The components needed for the design of spoken dialogue systems (SDSs)
that can communicate in a realistic human fashion have seen rapid
advancements in recent years (e.g. \citet{li_persona-based_2016,zhou_design_2018,skerry-ryan_towards_2018}).
However, an element of natural spoken conversation that is often overlooked
in SDS design is the timing of system responses. Many turn-taking
components for SDSs are designed with the objective of avoiding interrupting
the user while keeping the lengths of gaps and overlaps as low as
possible e.g. \citet{raux_finite-state_2009}. This approach does
not emulate naturalistic response offsets, since in human-human conversation
the distributions of response timing offsets have been shown to differ
based on the context of the first speaker's turn and the context of
the addressee's response \citep{sacks_simplest_1974,levinson_timing_2015,heeman_turn-taking_2017}. 
It has also been shown that listeners have different anticipations 
about upcoming responses based on the length of a silence before a response \citep{bogels_conversational_2019}.
If we wish to realistically generate offsets distributions in
SDSs, we need to design response timing models that take into
account the context of the user's speech and the upcoming system response.
For example, offsets where the first speaker's turn is a 
\emph{backchannel} occur in overlap more frequently
\citep{levinson_timing_2015}. It has also been observed that \emph{dispreferred
} responses (responses that are not in line with the suggested action
in the prior turn) are associated
with longer delays \citep{kendrick_timing_2015,bogels_conversational_2019}.

\begin{figure}[t!]
\centering
\includegraphics[width=0.99\linewidth]{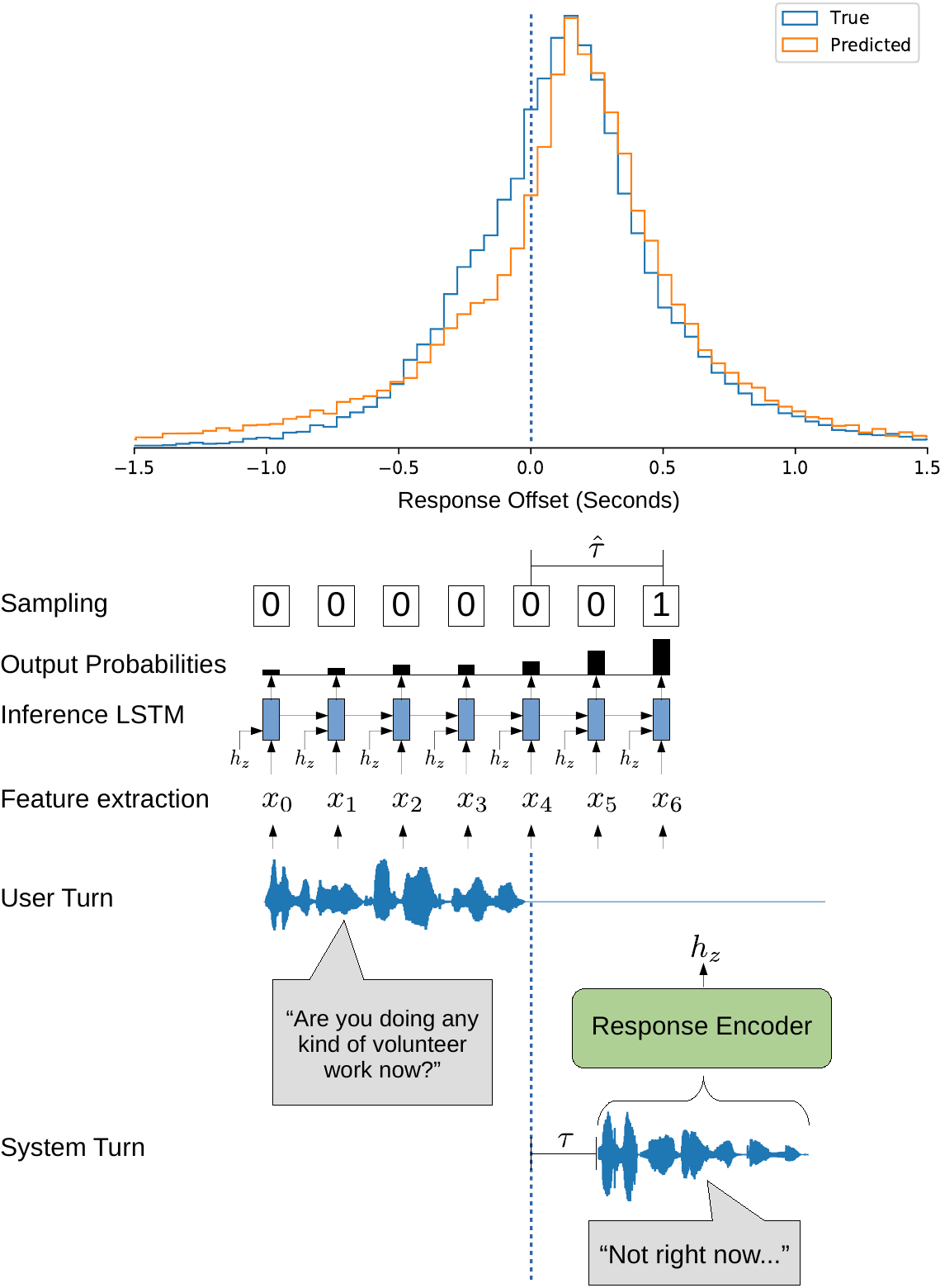}
\caption{Overview of how our model generates the distribution of turn-switch offset timings using an encoding of a dialogue system response $h_{z}$, and features extracted from the user's speech $x_{n}$. \label{fig:Overview}}
\end{figure}

\paragraph{Overview\label{paragraph:overview}}

We propose a neural model for generating these response timings in
SDSs (shown in Fig. \ref{fig:Overview}). 
The response timing network (RTNet) operates
using both acoustic and linguistic features extracted from user and
system turns. The two main components are an encoder, which encodes
the system response $h_{z}$, and an inference network, which takes
a concatenation of user features ($x_{n}$) and $h_{z}$.  RTNet operates within an incremental SDS framework 
\citep{schlangen_general_2011} where information
about upcoming system responses may be available before the user has
finished speaking. RTNet also functions independently of higher-level
turn-taking decisions that are traditionally made by the dialogue
manager (DM) component. Typically, the DM decides when the system
should take a turn and also supplies the natural language generation (NLG)
component with a semantic representation of the system response (e.g.
intents, dialogue acts, or an equivalent neural representation). 
Any of the system
response representations that are downstream from the DM's output
representation (e.g. lexical or acoustic features) can potentially
be used to generate the response encoding. Therefore, 
we assume that the decision for the system to
take a turn has already been made by the DM and our objective is to
predict (on a frame-by-frame basis) the appropriate time to trigger the
system turn.  

It may be impractical in an incremental framework to generate
a full system response and then re-encode it using the response encoder
of RTNet. To address this issue, we propose an extension of RTNet that
uses a variational autoencoder (VAE) \citep{kingma_auto-encoding_2014}
to train an interpretable latent space which can be used to bypass
the encoding process at inference-time. This extension (RTNet-VAE) allows the benefit of having a data-driven neural
representation of response encodings that can be manipulated without
the overhead of the encoding process. This representation can be manipulated
using vector algebra in a flexible manner by the DM to generate appropriate
timings for a given response.

Our model's architecture is similar to VAEs with recurrent encoders
and decoders proposed in \citet{bowman_generating_2016,ha_neural_2018,roberts_hierarchical_2018}.
Our use of a VAE to cluster dialogue acts is similar to the approach
used in \citet{zhao_learning_2017}. 
Our vector-based representation
of dialogue acts takes inspiration from the `attribute vectors'
used in \citet{roberts_hierarchical_2018} for learning musical structure
representations.
Our model is also related to continuous turn-taking systems
\citep{skantze_towards_2017} in that our model is trained
to predict future speech behavior on a frame-by-frame basis. 
The encoder uses a multiscale RNN architecture similar to the one proposed in \citet{roddy_multimodal_2018} to fuse information across modalities.
Models that intentionally generate responsive overlap have been proposed
in \citet{devault_incremental_2011,dethlefs_optimising_2012}. While
other models have also been proposed that generate appropriate response
timings for fillers \citep{nakanishi_generating_2018,lala_analysis_2019} and backchannels \citep{morency_probabilistic_2010, meena_data-driven_2014, lala_attentive_2017}.

This paper is structured as follows: First, we present how our 
dataset is structured and our training objective. Then, in sections 
\ref{subsec:RTNet} and \ref{subsec:RTNet-VAE} we present details of our two models, 
RTNet and RTNet-VAE. Section \ref{input-features} presents our input feature representations.
In section \ref{training-testing} we discuss our training and testing procedures. In sections \ref{subsec:RTNet-Discussion} and \ref{subsection:RTNet-VAE-Discussion} 
we analyze the performance of both RTNet and RTNet-VAE. 
Finally, in section \ref{sec:Listening-Tests} 
we present the results of a human listener 
test.

\section{Methodology}
\paragraph{Dataset}

\begin{figure*}[ht]
\begin{centering}
\includegraphics[width=\linewidth]{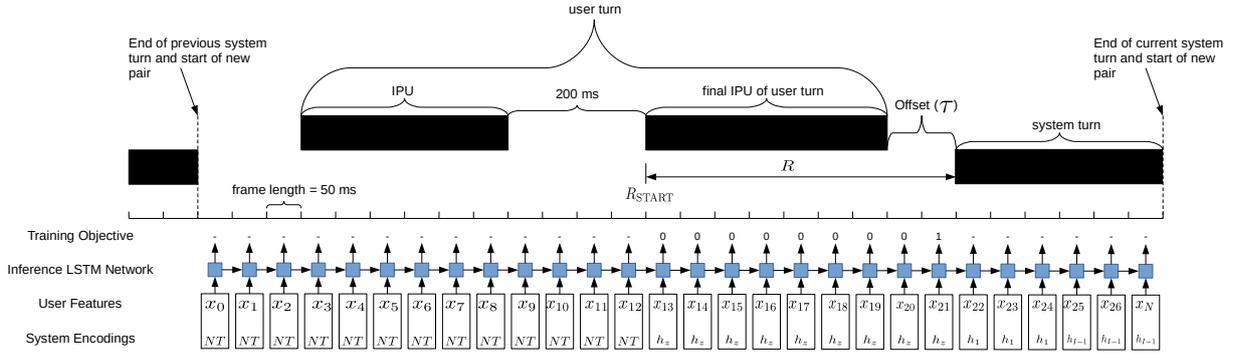}
\par\end{centering}
\caption{Segmentation of data into \emph{turn pairs},
and how the inference LSTM makes predictions. \label{fig:Inference}}
\end{figure*}

Our dataset is extracted from the Switchboard-1 Release 2 corpus \citep{godfrey_switchboard-1_1997}.
Switchboard has 2438 dyadic telephone conversations with a
total length of approximately 260 hours. 
The dataset consists of pairs of adjacent turns by different speakers 
which we refer to as 
\emph{turn pairs} (shown in Fig. \ref{fig:Inference}). 
Turn pairs are automatically extracted from orthographic annotations
using the following procedure: We extract frame-based speech-activity
labels for each speaker using a frame step-size of 50ms. 
The frame-based representation
is used to partition each person's speech signal into \emph{interpausal
units} (IPUs). We define IPUs as segments of speech by a person
that are separated by pauses of 200ms or greater. IPUs are then
used to automatically extract \emph{turns}, which we define as consecutive
IPUs by a speaker in which there is no speech by the other speaker
in the silence between the IPUs. 
A turn pair is then defined as being any two adjacent turns by different speakers. 
The earlier of the two turns
in a pair is considered to be the \emph{user turn} and the second
is considered to be the \emph{system turn}. 

\paragraph{Training Objective\label{par:Training-Objective}}

Our training objective is to predict the start of the system
turn one frame ahead of the ground truth start time. The target labels
in each turn pair are derived from the ground truth speech activity
labels as shown in Fig. \ref{fig:Inference}. Each 50 ms frame has a
label $y\in\{0,1\}$, which consists of the ground truth voice activity
shifted to the left by one frame. As shown in the figure, we only
include frames in the span $R$ in our training loss. We define the
span $R$ as the frames from the beginning of the last IPU in the
user turn to the frame immediately prior to the start of the system
turn. 

We do not predict at earlier frames since we assume that at these
mid-turn-pauses the DM has not decided to take
a turn yet, either because it expects the user to continue, or it
has not formulated one yet. As mentioned previously in section \ref{paragraph:overview}, we design
RTNet to be abstracted from the turn-taking
decisions themselves. If we were to include pauses prior to the turn-final
silence, our response generation system would be additionally burdened
with making turn-taking decisions, namely, classifying between mid-turn-pauses
and end-of-turn silences. We therefore make the modelling assumption
that the system's response is formulated at some point during the
user's turn-final IPU. To simulate this assumption we sample an index
$R_{\textrm{START}}$ from the span of $R$ using a uniform distribution.
We then use the reduced set of frames from $R_{\textrm{START}}$ to
$R_{\textrm{END}}$ in the calculation of our loss. 

\subsection{Response Timing Network (RTNet)\label{subsec:RTNet}}

\paragraph{Encoder}

\begin{figure*}
\begin{centering}
\includegraphics[width=0.7\linewidth]{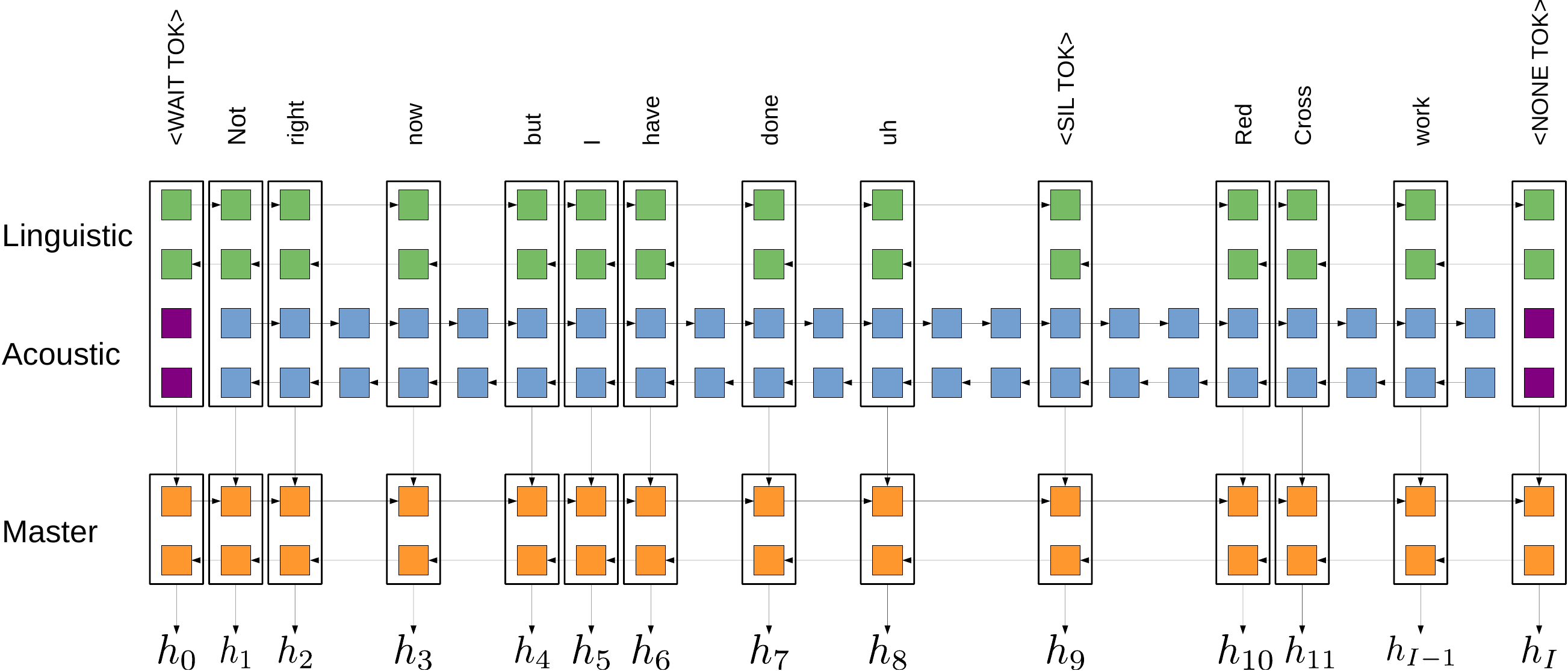}
\par\end{centering}
\caption{The encoder is three stacked Bi-LSTMs. We use special embeddings (shown in purple) to represent the acoustic states
corresponding to the first and last tokens (WAIT and NONE) of
the system's turn.  \label{fig:Response-Encoder}}
\end{figure*}

The encoder of RTNet (shown in Fig. \ref{fig:Response-Encoder}) fuses
the acoustic and linguistic modalities from a system response using
three bi-directional LSTMs. Each modality is processed at independent
timescales and then fused in a master Bi-LSTM which operates at the
linguistic temporal rate. The output of the master Bi-LSTM is a sequence
of encodings $h_{0},h_{1},...h_{I}$, where each encoding is a concatenation
of the forward and backward hidden states of the master Bi-LSTM at each word
index. 

The linguistic Bi-LSTM takes as input the sequence of 300-dimensional
embeddings of the tokenized system response. We use three special
tokens: SIL, WAIT, and NONE. The SIL token is used whenever there
is a gap between words that is greater than the frame-size (50ms).
The WAIT and NONE tokens are inserted as the first and last tokens of the system
response sequence respectively. The concatenation $[h_{0};h_{1};h_{I}]$ 
is passed as input to a RELU
layer (we refer to this layer as the \emph{reduction layer}) which outputs
the $h_{z}$ encoding. 
The $h_{z}$ encoding is used (along with user
features) in the concatenated input to the inference network. Since
the WAIT embedding corresponds to the $h_{0}$ output of the master
Bi-LSTM and the NONE embedding corresponds to $h_{I}$, the two embeddings
serve as ``triggering'' symbols that allow the linguistic and master
Bi-LSTM to output relevant information accumulated in their cell states.

The acoustic Bi-LSTM takes as input the sequence of acoustic features
and outputs a sequence of hidden states at every 50ms frame. As shown
in Fig. \ref{fig:Response-Encoder}, we select the acoustic hidden
states that correspond to the starting frame of each linguistic token
and concatenate them with the linguistic hidden states. Since there
are no acoustic features available for the WAIT and NONE tokens, we
train two embeddings to replace these acoustic LSTM states (shown in purple in Fig. \ref{fig:Response-Encoder}). The use of acoustic embeddings results in there being no connection between the WAIT acoustic embedding and the first acoustic hidden state. For this reason we include $h_{1}$ in the $[h_{0};h_{1};h_{I}]$ concatenation, in order to make it easier for information captured by the the acoustic bi-LSTM to be passed through to the final concatenation. 

\paragraph{Inference Network}

The aim of our inference network is to predict a sequence of output
probabilities $Y=[y_{R_{{\rm START}}},y_{R_{{\rm START}}+1},...,y_{N}]$
using a response encoding $h_{z}$, and a sequence of user
features $X=[x_{0},x_{1},...,x_{N}]$. We use a 
a single-layer LSTM (shown in Fig. \ref{fig:Inference}) 
which is followed by a sigmoid layer to produce
the output probabilities:
\[
[h_{n};c_{n}]=\mathrm{LSTM_{inf}}([x_{n};h_{z}],[h_{n-1};c_{n-1}])
\]
\[
y_{n}=\boldsymbol{\sigma}(\boldsymbol{W}_{h}h_{n}+\boldsymbol{b}_{h})
\]
Since there are only two possible output values in a generated sequence
\{0,1\}, and the sequence ends once we predict 1, the inference
network can be considered an autoregressive model where 0 is passed
implicitly to the subsequent time-step. To generate an output sequence,
we can sample from the distribution $p(y_{n}=1|y_{R_{{\rm START}}}=0,y_{R_{{\rm START}}+1}=0,...,y_{n-1}=0,X_{0:n},h_{z})$
using a Bernoulli random trial at each time-step. For frames prior
to $R_{{\rm START}}$ the output probability is fixed to 0, since $R_{{\rm START}}$
is the point where the DM has formulated the response. During
training we minimize the binary cross entropy loss ($L_{{\rm BCE}}$)
between our ground truth objective and our output predictions $Y$.

\subsection{RTNet-VAE \label{subsec:RTNet-VAE}}

\paragraph{Motivation \label{par:VAE-Motivation}}

\begin{figure}
\begin{centering}
\includegraphics[width=0.7\linewidth]{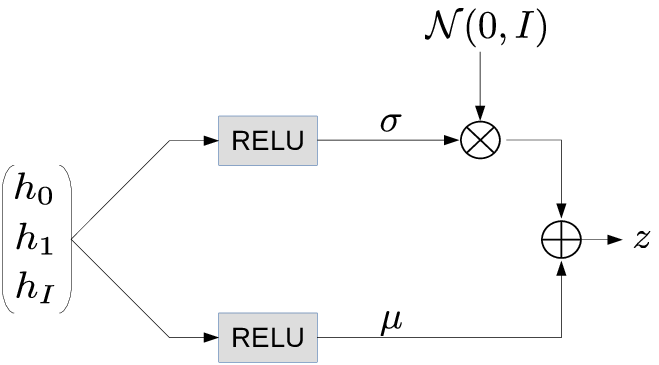}
\par\end{centering}
\caption{VAE \label{fig:VAE}}
\end{figure}

A limitation of RTNet is that it may be impractical 
to encode system turns before triggering a response.
For example, if we wish to apply RTNet using generated system
responses, at run-time the RTNet component would have to wait for
the full response to be generated by the NLG, which would result in
a computational bottleneck. If the NLG system is incremental, it may
also be desirable for the system to start speaking before the entirety
of the system response has been generated.

\paragraph{VAE}

To address this, we bypass the encoding stage by directly
using the semantic representation output from the DM to control the
response timing encodings. We do this by replacing the reduction layer
with a VAE (Fig. \ref{fig:VAE}). To train the VAE, we use the same
concatenation of encoder hidden states as in the RTNet reduction
layer ($[h_{0};h_{1};h_{I}]$). We use a dimensionality reduction
RELU layer to calculate $h_{{\rm reduce}}$, which is then split into
$\mu$ and $\hat{\sigma}$ components
via two more RELU layers. $\hat{\sigma}$ is passed through an exponential
function to produce $\sigma$, a non-negative standard deviation parameter.
We sample the latent variable $z$ with the standard VAE
method using $\mu$, $\sigma$, and a random vector from the standard
normal distribution $\mathcal{N}(0,\mathbf{I})$. A dimensionality expansion
RELU layer is used to transform $z$ into the response encoding $h_{z}$,
which is the same dimensionality as the output of the encoder:
\[
h_{{\rm reduce}}={\rm RELU}(W_{{\rm reduce}}[h_{0};h_{1};h_{I}]+b_{{\rm reduce}})
\]
\[
\mu={\rm RELU}(W_{\mu}h_{{\rm reduce}}+b_{\mu})
\]
\[
\hat{\sigma}={\rm RELU}(W_{\sigma}h_{{\rm reduce}}+b_{\sigma})
\]
\[
\sigma=\exp(\frac{\hat{\sigma}}{2})
\]
\[
z=\mu+\sigma\odot\mathcal{N}(0,\mathbf{I})
\]
\[
h_{z}={\rm RELU}(W_{{\rm expand}}z+b_{{\rm expand}})
\]
We impose a Gaussian prior over the latent space using a Kullback-Liebler
(KL) divergence loss term:
\[
L_{{\rm KL}}=-\frac{1}{2N_{z}}(1+\hat{\sigma}-\mu^{2}-\exp(\hat{\sigma}))
\]
The $L_{{\rm KL}}$ loss measures the distance of the generated distribution
from a Gaussian with zero mean and unit variance.  $L_{{\rm KL}}$
is combined with $L_{{\rm BCE}}$ using a weighted sum:
\[
L=L_{{\rm BCE}}+w_{{\rm KL}}L_{{\rm KL}}
\]
As we increase the value of $w_{{\rm KL}}$ we increasingly enforce
the Gaussian prior on the latent space. In doing so our aim is to
learn a smooth latent space in which similar types of responses are
organized in similar areas of the space. 

\paragraph{Latent Space \label{par:Latent-Space}}

During inference we can skip the encoding stage of RTNet-VAE and sample
$z$ directly from the latent space on the basis of the input semantic
representation from the dialogue manager. Our sampling approach is
to approximate the distribution of latent variables for a given response-type
using Gaussians. For example, if we have a collection of labelled
\emph{backchannel }responses (and their corresponding $z$ encodings)
we can approximate the distribution of $p(z|{\rm label=}\mathit{backchannel})$
using an isotropic Gaussian by simply calculating $\mu_{{\it backchannel}}$
and $\sigma_{{\it backchannel}}$, the maximum likelihood mean and
standard deviations of each of the $z$ dimensions. These vectors
can also be used to calculate directions
in the latent space with different semantic characteristics and
then interpolate between them. 
\subsection{Input Feature Representations\label{input-features}}
\begin{figure*}
\begin{centering}
\includegraphics[width=0.8\linewidth]{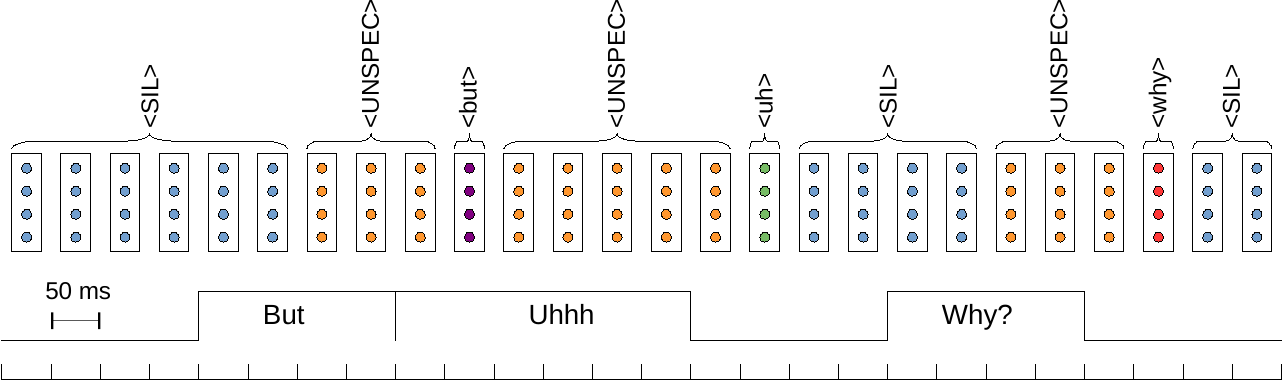}
\par\end{centering}
\caption{The user's linguistic feature representation scheme. The embedding
for each word is triggered 100 ms after the ground truth end of the
word, to simulate ASR delay. The UNSPEC embedding begins 100ms after a word's start frame and holds information about whether a word is being spoken (before
it has been recognized) and the length of each word. \label{fig:Unspec}}
\end{figure*}

\paragraph{Linguistic Features}
We use the word annotations from the ms-state
transcriptions as linguistic features. These annotations give us the timing for the starts and ends of all words in the corpus. As our feature representation,
we use 300 dimensional word embeddings that are initialized with GloVe
vectors \citep{pennington_glove_2014} and then jointly optimized
with the rest of the network. In total there are
30080 unique words in the annotations. We reduced the embedding number
down to $10000$ by merging embeddings that had low word counts
with the closest neighbouring embedding (calculated using cosine distance).

We also introduce four additional tokens that are specific to our
task: SIL, WAIT, NONE, and UNSPEC. SIL is used whenever there is a silence.
WAIT and NONE are used at the start and end of all the system encodings, respectively. 
The use of UNSPEC (unspecified) is shown in
Fig. \ref{fig:Unspec}. UNSPEC was introduced to represent temporal information
in the linguistic embeddings. We approximate the processing delay in ASR
by delaying the annotation by 100 ms after the ground truth frame
where the user's word ended. This 100 ms delay was proposed in \citet{skantze_towards_2017} as a necessary assumption to modelling linguistic features in offline continuous systems.
However, since voice activity detection
(VAD) can supply an estimate of when a word has started, we 
propose that we can
use this information to supply the network with the UNSPEC embedding
100ms after the word has started. 

\paragraph{Acoustic Features}

We combine 40 log-mel filterbanks,
and 17 features from the GeMAPs feature set \citep{eyben_geneva_2016}.
The GeMAPs features are the complete set excluding the
MFCCs (e.g. pitch, intensity, spectral flux, jitter, etc.). Acoustic features were extracted using a 50ms framestep. 

\subsection{Experimental Settings}

\paragraph{Training and Testing Procedures\label{training-testing}}
The training, validation, and test sets consist of 1646, 150, 642 conversations
respectively with 151595, 13910, and 58783 turn pairs. The test set
includes all of the conversations from the NXT-format annotations
\citep{calhoun_nxt-format_2010}, which include references to the Switchboard Dialog Act Corpus (SWDA) \citep{stolcke_dialogue_2000} annotations.
We include the entirety of the NXT annotations in our test set
so that we have enough labelled dialogue act samples to analyse
the distributions. 

We used the following hyperparameter settings in our experiments:
The inference, acoustic, linguistic, and master LSTMs each had 
hidden sizes of 1024, 256, 256, and 512 (respectively).
We used a latent variable size of 4, a batch size of 128, and L2 regularization of 1e-05. 
We used the Adam optimizer with an initial learning rate of 5e-04. We trained each model for 15000 iterations, with learning rate reductions by a factor of 0.1 after 9000, 11000, 13000, and 14000 iterations.

While we found that randomizing $R_{\textrm{START}}$ during training
was important for the reasons given in Section \ref{par:Training-Objective},
it presented issues for the stability and reproducibility of our evaluation
and test results for $L_{{\rm BCE}}$ and $L_{{\rm KL}}$. We therefore randomize during training and sampling, but when calculating the test losses (reported in Table
\ref{tab:Results-table}) we fix $R_{\textrm{START}}$ to be the first
frame of the user's turn-final IPU. 

We also calculate the mean absolute error (MAE), given in seconds, from the ground truth response offsets to the generated output offsets. When sampling for the calculation of MAE, it is necessary to increase the length of the turn pair since the response time may
be triggered by the sampling process \emph{after} the ground truth
time. We therefore pad the user's features with 80 extra frames in which we 
simulate silence artificially using acoustic features. During sampling, we use the same $R_{\textrm{START}}$ randomization process that was used during training, rather than fixing it to the start of the user's turn-final IPU. For each model we perform the sampling procedure on the test set three times and report the mean error in Table \ref{tab:Results-table}.

\paragraph{Best Fixed Probability}

To the best of our knowledge, there aren't any other published models that 
we can directly compare ours to. However, we can calculate the best performance
that can be achieved using a fixed value for $y$. The best possible
fixed $y$ for a given turn pair is: $y_{{\rm tp}}=\frac{1}{(R_{\textrm{END}}-\text{\ensuremath{R_{\textrm{START}}}})/{\rm FrameLength}}$.
The best fixed $y$ for a set of turn pairs is given by the expected
value of $y_{{\rm tp}}$ in that set: $y_{{\rm fixed}}=\mathbb{E}[y_{{\rm tp}}]$.
This represents the best performance that we could achieve if we did
not have access to any user or system features. We can use the fixed
probability model to put the performance of the rest of our models
into context.

\begin{table}
\begin{centering}
\begin{tabular}{|c|c|c|c|c|c|}
\hline 
{\tiny{}\#} & {\tiny{}Model} & {\tiny{}$L_{{\rm BCE}}$} & {\tiny{}$L_{{\rm KL}}$} & {\tiny{}MAE} & {\tiny{}Details}\tabularnewline
\hline 
\hline 
{\tiny{}1} & {\tiny{}Full Model} & \textbf{\tiny{}0.1094} & {\tiny{}--} & \textbf{\tiny{}0.4539} & \multirow{1}{*}{{\tiny{}No VAE}}\tabularnewline
\hline 
{\tiny{}2} & {\tiny{}Fixed Probability} & {\tiny{}0.1295} & {\tiny{}--} & {\tiny{}1.4546} & {\tiny{}Fixed Probability}\tabularnewline
\hline 
{\tiny{}3} & {\tiny{}No Encoder} & {\tiny{}0.1183} & {\tiny{}--} & {\tiny{}0.4934} & \multirow{3}{*}{{\tiny{}Encoder Ablation}}\tabularnewline
\cline{1-5} \cline{2-5} \cline{3-5} \cline{4-5} \cline{5-5} 
{\tiny{}4} & {\tiny{}Only Acoustic} & {\tiny{}0.1114} & {\tiny{}--} & {\tiny{}0.4627} & \tabularnewline
\cline{1-5} \cline{2-5} \cline{3-5} \cline{4-5} \cline{5-5} 
{\tiny{}5} & {\tiny{}Only Linguistic} & {\tiny{}0.1144} & {\tiny{}--} & {\tiny{}0.4817} & \tabularnewline
\hline 
{\tiny{}6} & {\tiny{}Only Acoustic} & {\tiny{}0.1112} & {\tiny{}--} & {\tiny{}0.5053} & \multirow{2}{*}{{\tiny{}Inference Ablation}}\tabularnewline
\cline{1-5} \cline{2-5} \cline{3-5} \cline{4-5} \cline{5-5} 
{\tiny{}7} & {\tiny{}Only Linguistic} & {\tiny{}0.1167} & {\tiny{}--} & {\tiny{}0.4923} & \tabularnewline
\hline 
{\tiny{}8} & {\tiny{}$w_{KL}=0.0$} & {\tiny{}0.1114} & {\tiny{}3.3879} & {\tiny{}0.4601} & \multirow{5}{*}{{\tiny{}Inclusion of VAE }}\tabularnewline
\cline{1-5} \cline{2-5} \cline{3-5} \cline{4-5} \cline{5-5} 
{\tiny{}9} & {\tiny{}$w_{KL}=10^{-4}$} & {\tiny{}0.1122} & {\tiny{}1.5057} & {\tiny{}0.4689} & \tabularnewline
\cline{1-5} \cline{2-5} \cline{3-5} \cline{4-5} \cline{5-5} 
{\tiny{}10} & {\tiny{}$w_{KL}=10^{-3}$} & {\tiny{}0.1125} & {\tiny{}0.8015} & {\tiny{}0.4697} & \tabularnewline
\cline{1-5} \cline{2-5} \cline{3-5} \cline{4-5} \cline{5-5} 
{\tiny{}11} & {\tiny{}$w_{KL}=10^{-2}$} & {\tiny{}0.1181} & {\tiny{}0.0000} & {\tiny{}0.5035} & \tabularnewline
\cline{1-5} \cline{2-5} \cline{3-5} \cline{4-5} \cline{5-5} 
{\tiny{}12} & {\tiny{}$w_{KL}=10^{-1}$} & {\tiny{}0.1189} & \textbf{\tiny{}0.0000} & {\tiny{}0.5052} & \tabularnewline
\hline 
\end{tabular}
\par\end{centering}
\caption{Experimental results on our test set. Lower is better in all cases.
Best results shown in bold. \label{tab:Results-table}}
\end{table}

\begin{figure}[ht!]
    \centering
    \begin{subfigure}[b]{0.23\textwidth}
    \centering
        \includegraphics[width=1.0\textwidth]{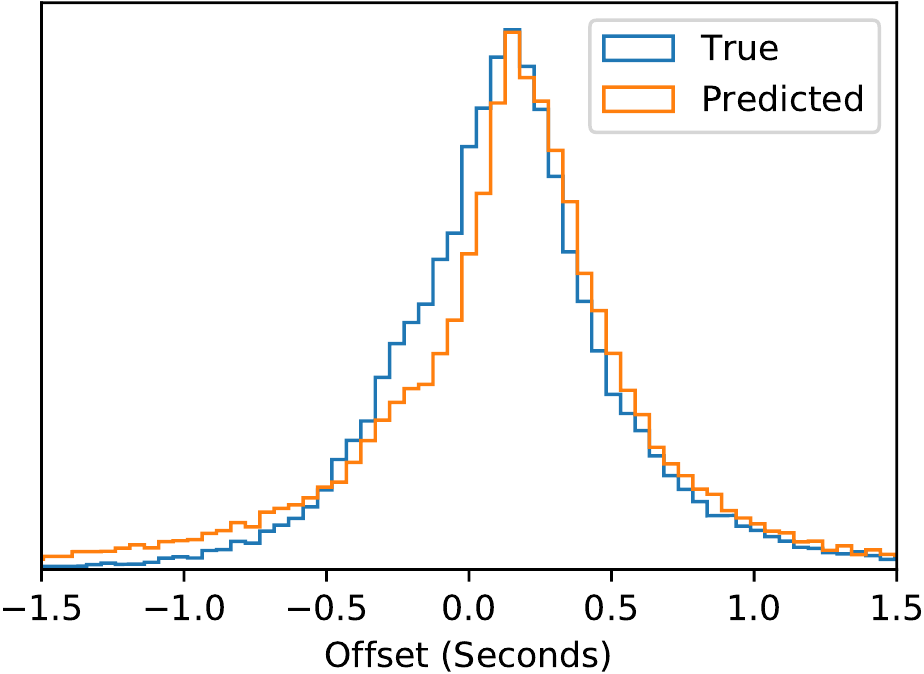}
        \caption{Full Model}
        \label{fig:full-model}
    \end{subfigure}
    \hfill{}
    \begin{subfigure}[b]{0.23\textwidth}
    \centering
        \includegraphics[width=1.0\textwidth]{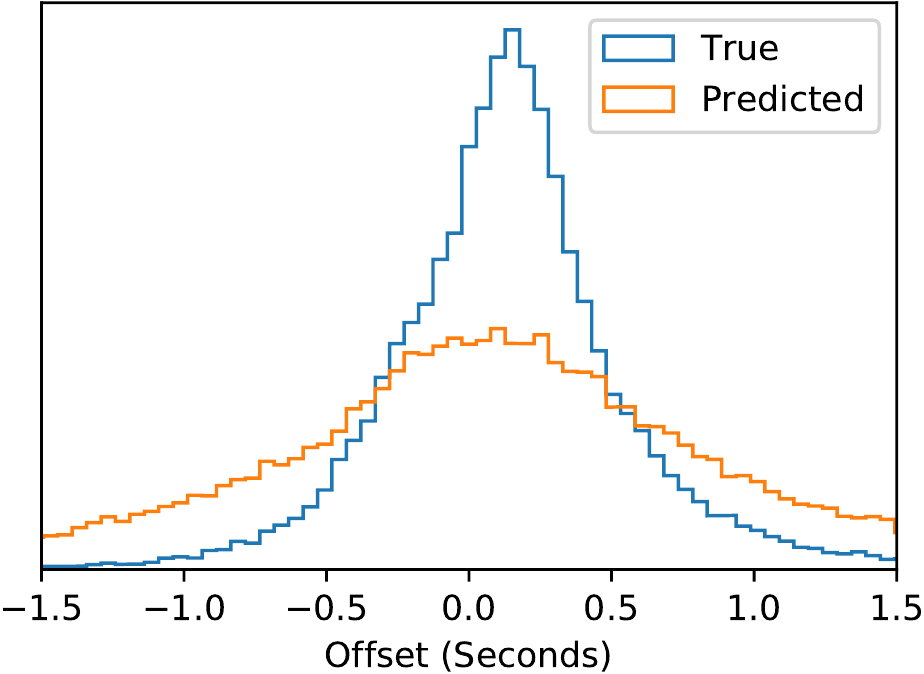}
        \caption{Fixed Probability}
        \label{fig:fixed-prob}
    \end{subfigure}
    \caption{Generated offset distributions for the test set using the full model and the fixed probability (random) model.}\label{fig:offset_distributions}
\end{figure}
\section{Discussion}
\subsection{RTNet Discussion \label{subsec:RTNet-Discussion}}
\paragraph{RTNet Performance \label{paragraph:Full-Model-Performance}}

\begin{figure}
    \centering
    \begin{subfigure}[b]{0.23\textwidth}
    \centering
        \includegraphics[width=0.75\textwidth]{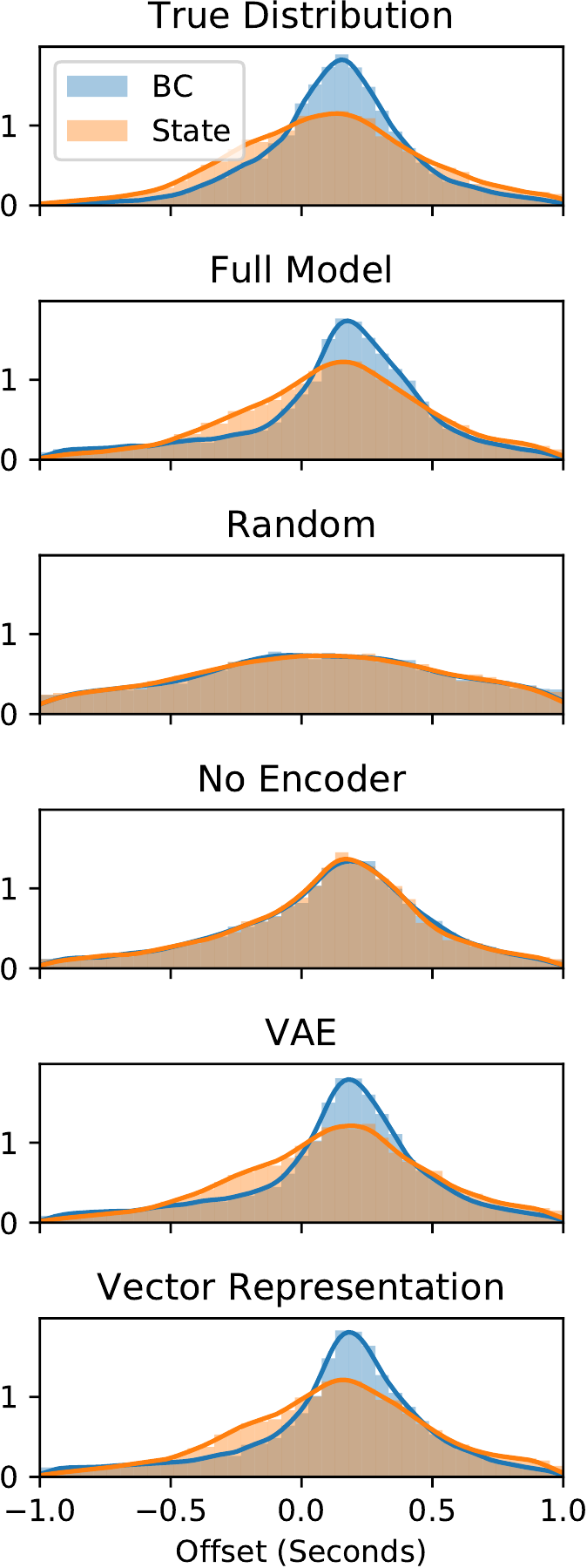}
        \caption{BC/Statement}
        \label{fig:bc-statement}
    \end{subfigure}
    \hfill{}
    \begin{subfigure}[b]{0.23\textwidth}
    \centering
        \includegraphics[width=0.75\textwidth]{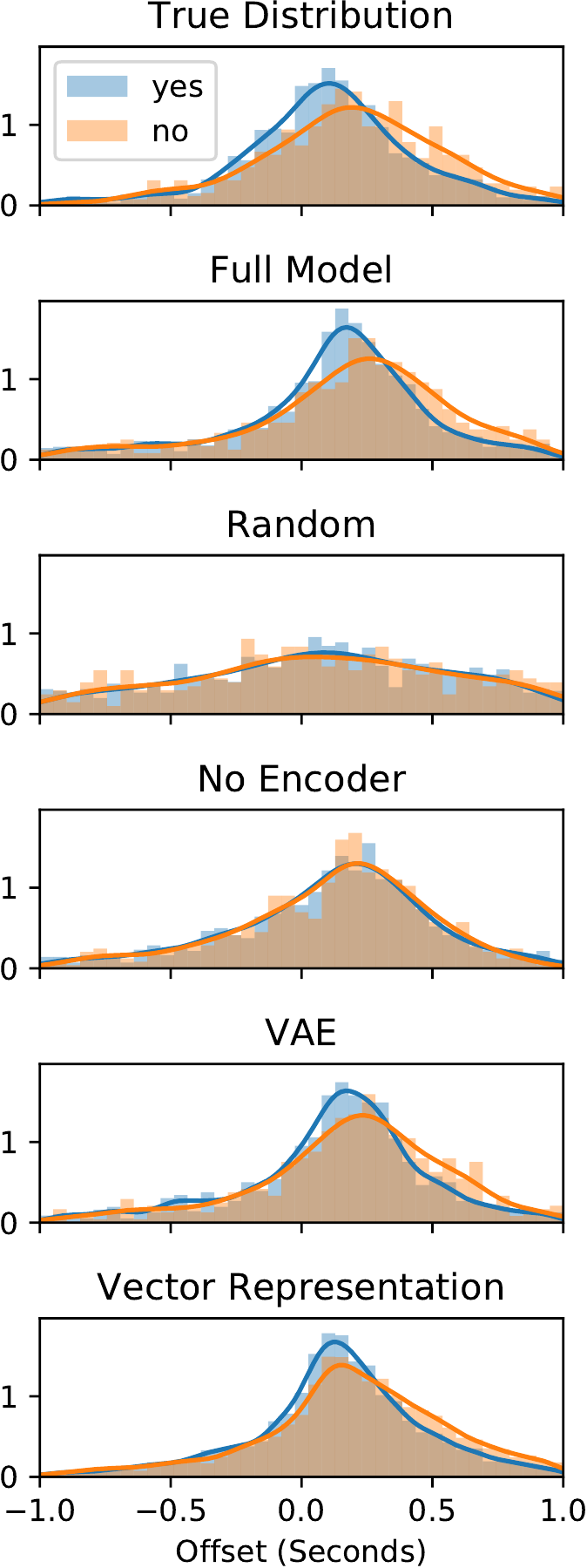}
        \caption{Yes/No}
        \label{fig:yes-no}
    \end{subfigure}
    \caption{Generated offset distributions for selected response dialogue acts using different model conditions.}\label{fig:offset_distributions2}
\end{figure}

The offset distribution for the full RTNet model is shown in Fig.
\ref{fig:full-model}. This baseline RTNet model is better able to replicate many of
the features of the true distribution in comparison with predicted offsets using the best possible fixed probability shown in Fig. \ref{fig:fixed-prob}.
The differences between the
baseline and the fixed probability distributions are reflected in
the results of rows 1 and 2 in Table \ref{tab:Results-table}.
In Fig. \ref{fig:full-model}, the model has the most trouble reproducing
the distribution of offsets between -500 ms and 0 ms. This part of
the distribution is the most demanding because it requires that
the model anticipate the user's turn-ending. From the plots it is
clear that our model is able to do that to a large degree. We observe that
after the user has stopped speaking (from 0 seconds onward) the generated
distribution follows the true distribution closely.

To look in more detail at how the system models the offset distribution 
we can investigate the generated distributions of labelled
response dialogue acts in our test set. Fig. \ref{fig:offset_distributions2} shows plots of \emph{backchannels} vs. \emph{statements} (Fig. \ref{fig:bc-statement}), and \emph{yes} vs. \emph{no} (Fig.\ref{fig:yes-no}) responses. In the second rows, we can see that the full model is
able to accurately capture the differences in the contours of the
true distributions. For example, in the \emph{no} dialogue acts, the
full model accurately generates a mode that is delayed (relative to
\emph{yes} dialogue acts).

\paragraph{Encoder Ablation \label{subsec:Encoder-Performance}}

The performance of the response encoder was analysed in an ablation
study, with results in rows 3 through 5 of Table
\ref{tab:Results-table}. Without the response encoder, there is a large decrease in performance, relative to the full model.
From looking at the encoders with only acoustic and linguistic modalities,
we can see that the results benefit more from the acoustic modality
than the linguistic modality. If we consider the impact of the encoder
in more detail, we would expect that the network would not be able
to model distributional differences between different types of DA
responses without an encoder. This is confirmed in the fourth rows
of Fig. \ref{fig:offset_distributions2}, where we show the generated distributions without the encoder. We can see that without the encoder, the distributions of the all
of the dialogue act offsets are almost exactly the same. 

\paragraph{Inference Network Ablation\label{paragraph:Inference-Network-Performance}}

\begin{figure}[ht!]
    \centering
    \begin{subfigure}[b]{0.23\textwidth}
    \centering
        \includegraphics[width=1.0\textwidth]{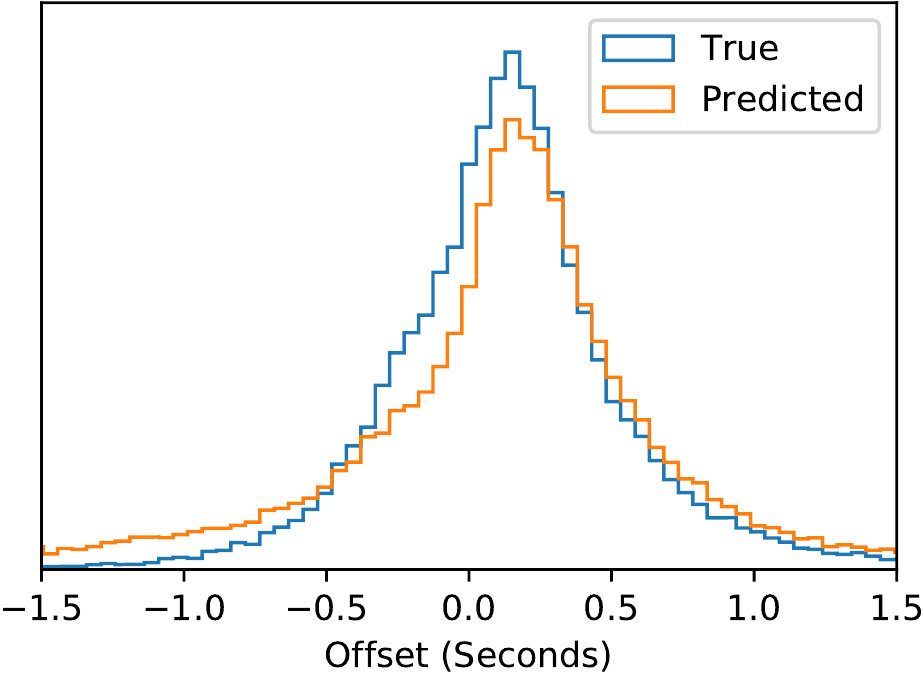}
        \caption{Only Acoustic}
        \label{fig:acous_abl}
    \end{subfigure}
    \hfill{}
    \begin{subfigure}[b]{0.23\textwidth}
    \centering
        \includegraphics[width=1.0\textwidth]{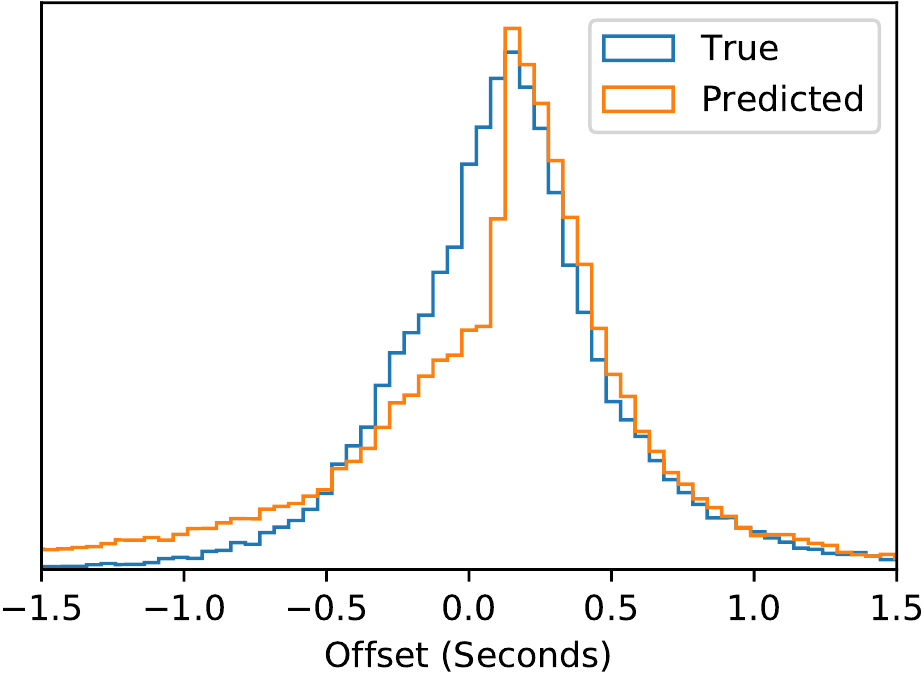}
        \caption{Only Linguistic}
        \label{fig:ling_abl}
    \end{subfigure}
    \caption{Generated offset distributions for the inference network ablation.}\label{fig:offset_distributions_ablation}
\end{figure}

In rows 6 and 7 of Table \ref{tab:Results-table} we present an ablation of the inference network. We can see that removing either the acoustic or linguistic features from the user's features is detrimental to the results. An interesting irregularity is observed in the results for the model that uses only acoustic features (row 6): the MAE is unusually high, relative to the $L_{\mathrm{BCE}}$. In all other rows, lower $L_{\mathrm{BCE}}$ corresponds to lower MAE. However, row 6 has the second lowest $L_{\mathrm{BCE}}$, while also having the second highest MAE. 

In order to examine this irregularity in more detail, we look at the generated distributions from the inference ablation, shown in Fig. \ref{fig:offset_distributions_ablation}. We observe that the linguistic features are better for predicting the mode of the distribution whereas the acoustic features are better at modelling the -100 ms to +150 ms region directly preceding the mode. Since word embeddings are triggered 100 ms after the end of the word, the linguistic features can be used to generate modal offsets in the 150 ms to 200 ms bin. We propose that, in the absence of linguistic features, there is more uncertainty about when the user's turn-end has occurred. Since the majority of all ground-truth offsets occur after the user has finished speaking, the unusually high MAE in row 6 could be attributed to this uncertainty in whether the user has finished speaking.

\subsection{RTNet-VAE Discussion \label{subsection:RTNet-VAE-Discussion}}

\begin{figure}
    \centering
    \begin{subfigure}[b]{0.23\textwidth}
    \centering
        \includegraphics[width=1.0\textwidth]{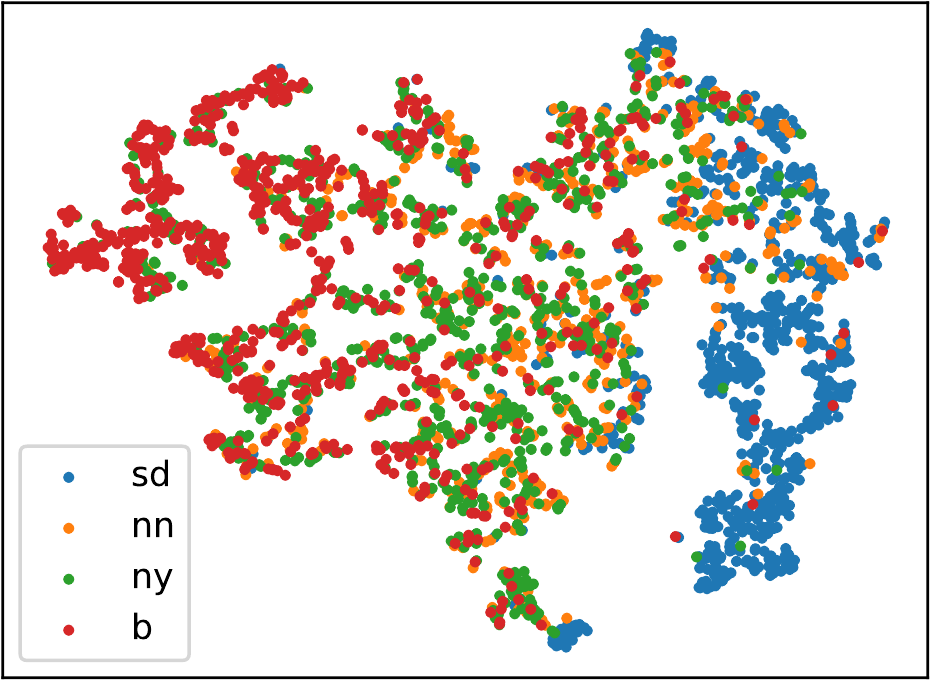}
        \caption{$w_{{\rm KL}}=0.0$}
        \label{fig:tsne-0}
    \end{subfigure}
    \hfill{}
    \begin{subfigure}[b]{0.23\textwidth}
    \centering
        \includegraphics[width=1.0\textwidth]{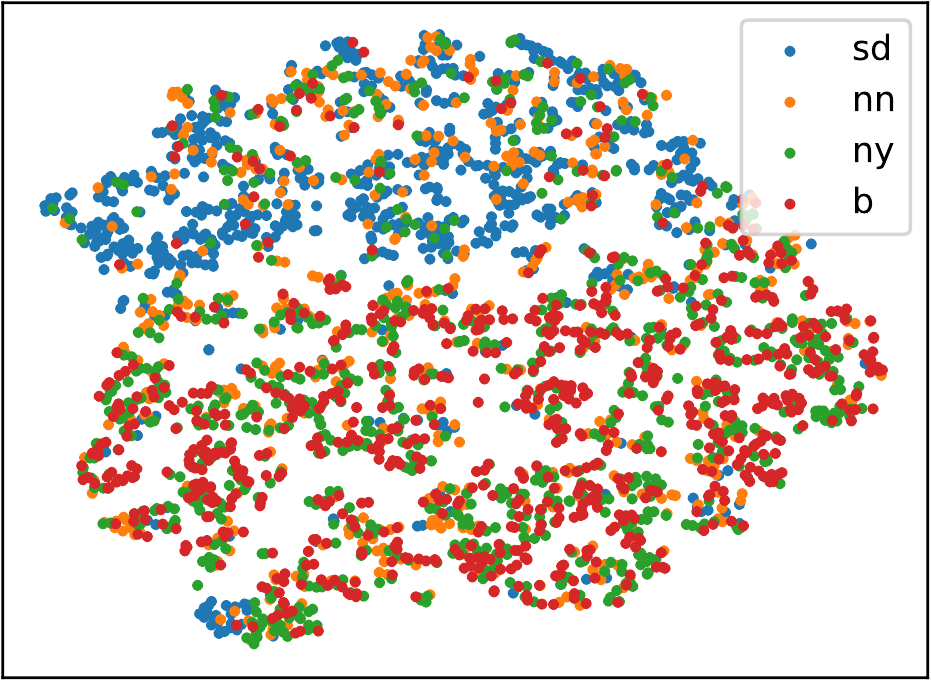}
        \caption{$w_{{\rm KL}}=10^{-3}$}
        \label{fig:tsne-0001}
    \end{subfigure}
    \caption{T-SNE plots of $z$ for four different dialogue acts using two different $w_{{\rm KL}}$ settings.}\label{fig:tsne}
\end{figure}

\begin{figure}
\begin{centering}
\includegraphics[width=0.99\linewidth]{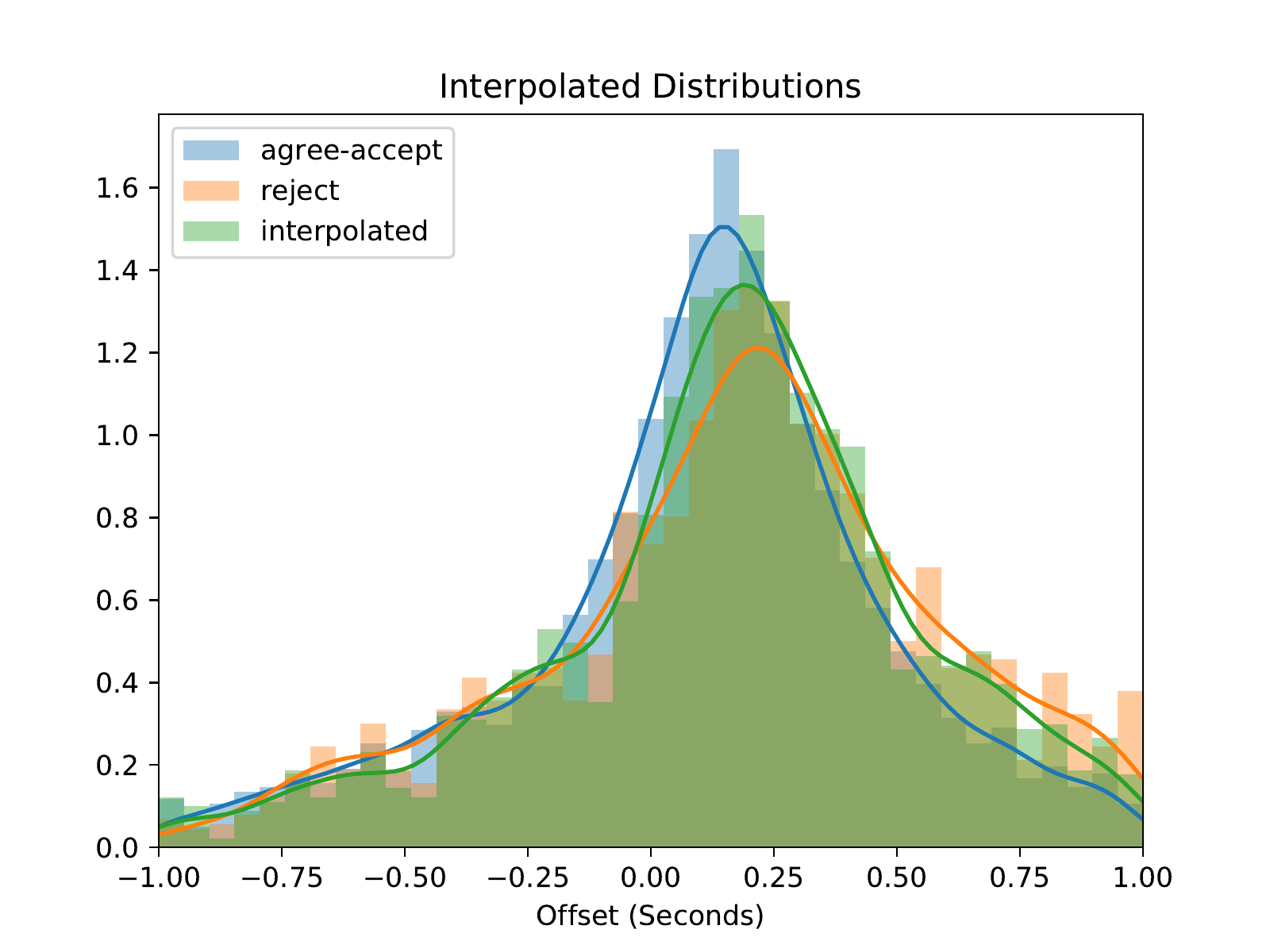}
\par\end{centering}
\caption{Interpolated distributions \label{fig:Interpolated-distributions}}
\end{figure}

\paragraph{RTNet-VAE Performance \label{paragraph:RTNet-VAE-Performance}}
In rows 8 through 12 of Table \ref{tab:Results-table} we show
the results of our experiments with RTNet-VAE with different settings of $w_{{\rm KL}}$. As $w_{{\rm KL}}$
is increased, the $L_{{\rm BCE}}$ loss increases while the $L_{{\rm KL}}$
loss decreases. Examining some example distributions of dialogue
acts generated by RTNet-VAE using $w_{{\rm KL}}=10^{-4}$ (shown in
the fifth rows of Fig. \ref{fig:offset_distributions2})
we can see that RTNet-VAE is capable of generating distributions that
are of a similar quality to those generated by RTNet (shown in the
second row). We also observe that RTNet-VAE
using $w_{{\rm KL}}=10^{-4}$ produces competitive results, in comparison to the full model.
These observations suggest that the inclusion of the VAE in pipeline
does not severely impact the overall performance. 

In Fig. \ref{fig:tsne} we show the latent variable $z$ generated
using RTNet-VAE and plotted using t-SNE \citep{maaten_visualizing_2008}. 
To show the benefits of imposing the
Gaussian prior, we show plots for with $w_{{\rm KL}}=0.0$ and $w_{{\rm KL}}=10^{-3}$.
The plots show the two-dimensional projection of four different types
of dialogue act responses: \emph{statements} (sd), \emph{no} (nn),
\emph{yes} (ny), and \emph{backchannels} (b). We can observe that
for both settings, the latent space is able to organize the responses
by dialogue act type, even though it is never explicitly trained on
dialogue act labels. For example, in both cases, statements (shown
in blue) are clustered at the opposite side of the distribution from
backchannels (shown in red). However, in the case of $w_{{\rm KL}}=0.0$
there are ``holes'' in the latent space. For practical applications 
such as interpolation of vector representations
of dialogue acts (discussed in the next paragraph), we would like a space that does
not contain any of these holes since they are less likely to have
semantically meaningful interpretations. When the Gaussian prior is
enforced (Fig. \ref{fig:tsne-0001}) we can see that the space is smooth and the distinctions
between dialogue acts is still maintained.

\paragraph{Latent Space Applications}

As mentioned in Section \ref{par:VAE-Motivation}, part of the appeal
in using the VAE in our model is that it enables us to discard
the response encoding stage. We can exploit the smoothness of the
latent space to skip the encoding stage by sampling directly from
the trained latent space. We can approximate the distribution of latent
variables for individual dialogue act response types using isotropic
Gaussians. This enables us to efficiently represent the dialogue acts
using mean and standard-deviation vectors, a pair for each dialogue
act. Fig. \ref{fig:offset_distributions2} shows
examples of distributions generated using Gaussian approximations
of the latent space distributions in the final rows. We can see that the generated outputs
have similar properties to the true distributions. 

We can use the same parameterized vector representations to interpolate
between different dialogue act parameters to achieve intermediate
distributions. This dimensional approach is flexible in that we give
the dialogue manager (DM) more control over the details of the distribution.
For example, if the objective of the SDS was to generate an \emph{agree}
dialogue act, we could control the \emph{degree} of agreement by interpolating
between \emph{disagree }and \emph{agree} vectors. Figure \ref{fig:Interpolated-distributions}
shows an example of a generated interpolated distribution. We can
see that the properties of the interpolated distribution (e.g. mode,
kurtosis) are perceptually ``in between'' the \emph{reject }and
\emph{accept} distributions.

\section{Listening Tests \label{sec:Listening-Tests}}

\begin{figure*}
    \centering
    \begin{subfigure}[b]{0.45\textwidth}
    \centering
        \includegraphics[width=1.0\textwidth]{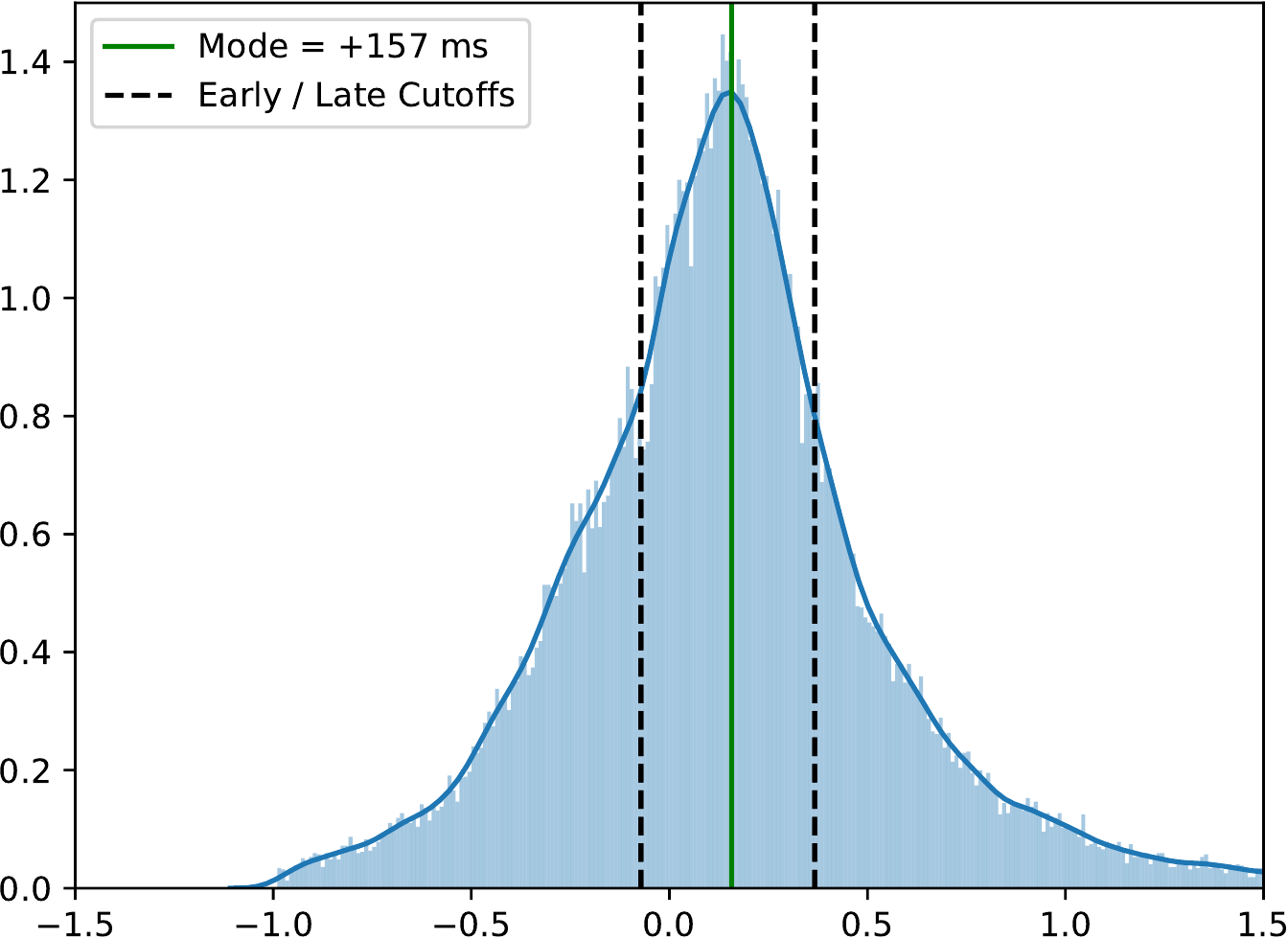}
        \caption{ \emph{Early}, \emph{modal}, and \emph{late} regions.}
        \label{fig:early_late_cuttoffs}
        \hfill{}
    \end{subfigure}
    \hfill{}
    \begin{subfigure}[b]{0.45\textwidth}
    \centering
        \includegraphics[width=0.9\textwidth]{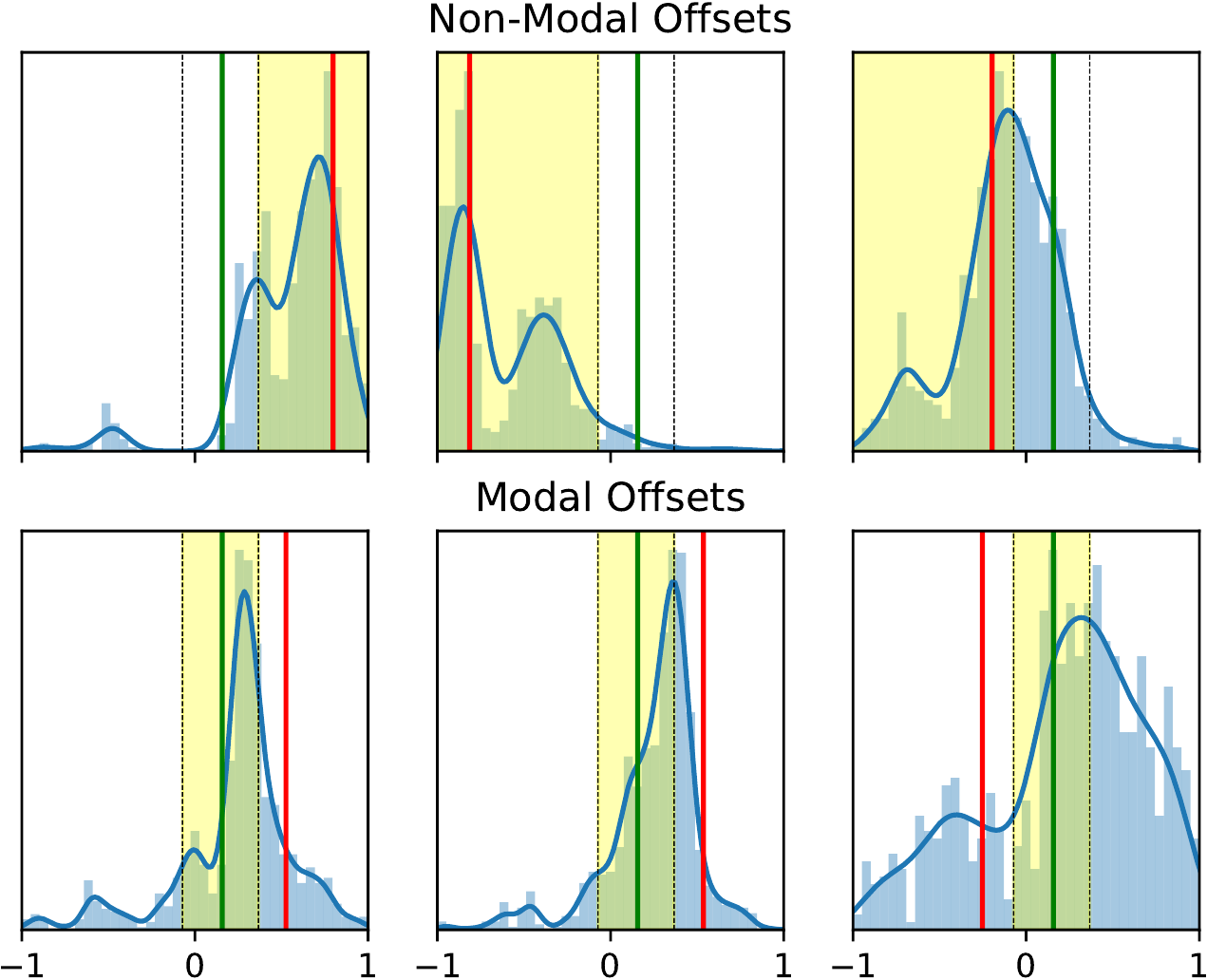}
        \caption{Generated distributions for six turn pairs.
        The highlighted regions indicate the region that was preferred by listeners. The red line indicates the ground truth offset.
        }
        \label{fig:individual_experiment}
    \end{subfigure}
    \caption{Listening test experiments}\label{fig:Listening-test}
\end{figure*}

It has shown that response timings vary based on the semantic content
of dialogue responses and the preceding turn \citep{levinson_timing_2015},
and that listeners are sensitive to these fluctuations in timing \citep{bogels_brain_2017}. 
However, the
question of whether certain response timings within different contexts
are considered more \emph{realistic} than others has not been fully
investigated. We design an online listening test to answer two questions:
(1) Given a preceding turn and a response, are some response timings
considered by listeners to be more realistic than others? (2) In cases
where listeners are sensitive to the response timing, is our model
more likely to generate responses that are considered realistic than
a system that generates a modal response time?

Participants were asked to make A/B choices
between two versions of a turn pair, where each version had a different
response offset. 
Participants were asked: "Which response timing sounds like it 
was produced in the real conversation?"
The turn pairs were drawn from our dataset and were limited to pairs
where the response was either \emph{dispreferred} or a \emph{backchannel}.
We limited the chosen pairs to those with ground truth offsets
that were either classified as \emph{early }or \emph{late}. We classified
offsets as early, modal, or late by segmenting the distribution of
all of the offsets in our dataset
into three partitions as shown in Fig. \ref{fig:early_late_cuttoffs}. 
The cutoff points for the
early and late offsets were estimated using a heuristic where we split
the offsets in our dataset into two groups at the mode of the distribution
(157 ms) and then used the median values of the 
upper (+367 ms) and lower (-72 ms) groups as the cutoff points. 
We selected eight examples of each dialogue act (four \emph{early} and four \emph{late}). 
We generated three different versions of each turn pair: \emph{true}, \emph{modal},
and \emph{opposite}. 
If the true offset was late, the opposite offset was
the mean of the early offsets (-316 ms). If the true offset was early,
the opposite offset was the mean of the late offsets (+760 ms). 

We had 25 participants (15 female, 10 male) who all wore headphones. We performed binomial tests
for the significance of a given choice in each question. For the questions
in the first half of the test, in which we compared \emph{true} vs.
\emph{opposite} offsets, 10 of the 16 comparisons were found to be
statistically significant ($p<0.05$). In all of the significant cases
the \emph{true} offset was was considered more realistic than the
\emph{opposite}. In reference to our first research question, this
result supports the conclusion that some responses are indeed considered
to be more realistic than others. For the questions in the second
half of the test, in which we compared \emph{true} vs. \emph{modal}
offsets, six out of the 16 comparisons were found to be statistically
significant. Of the six significant preferences, three were a preference
for the \emph{true} offset, and three were a preference for the \emph{modal}
offset. To investigate our second research question, we looked at
the offset distributions generated by our model for each of the six
significant preferences, shown in Fig. \ref{fig:individual_experiment}. For the turn pairs where
listeners preferred non-modal offsets (top row), the distributions
generated by our system deviate from the mode into the preferred area
(highlighted in yellow). In pairs where listeners preferred modal
offsets (bottom row) the generated distributions tend to have a mode
near the overall dataset mode (shown in the green line). We can conclude,
in reference to our second question, that in instances where listeners
are sensitive to response timings it is likely that our system will
generate response timings that are more realistic than a system that
simply generates the mode of the dataset.

\section{Conclusion}

In this paper, we have presented models that can be used to generate
the turn switch offset distributions of SDS system responses. It has
been shown in prior studies (e.g. \cite{bogels_conversational_2019})
that humans are sensitive to these timings and that they can impact
how responses are perceived by a listener. We would argue that they
are an important element of producing naturalistic interactions that
is often overlooked. With the advent of commercial SDS systems that
attempt to engage users over extended multi-turn interactions (e.g.
\cite{zhou_design_2018}) generating realistic response behaviors
is a potentially desirable addition to the overall experience.

\section*{Acknowledgments}

The ADAPT Centre for Digital Content Technology is funded under the SFI Research Centres Programme (Grant 13/RC/2106) and is co-funded under the European Regional Development Fund.

\bibliography{acl2020.bib}
\bibliographystyle{acl_natbib}

\end{document}